\documentclass{article}

\usepackage{microtype}
\usepackage{graphicx}
\usepackage{booktabs} 
\usepackage[table,dvipsnames]{xcolor}

\usepackage{hyperref}

\usepackage[accepted]{icml2025}

\usepackage{amsmath}
\usepackage{amssymb}
\usepackage{mathtools}
\usepackage{amsthm}
\usepackage[capitalize,noabbrev]{cleveref}
\usepackage{enumitem}

\usepackage{algorithm}
\usepackage{algpseudocode}
\usepackage[labelfont=bf]{caption}
\usepackage{subcaption}
\usepackage[utf8]{inputenc}
\usepackage{kotex}  

\theoremstyle{plain}

\theoremstyle{definition}

\theoremstyle{remark}

\usepackage[textsize=tiny]{todonotes}
\usepackage{multirow}

\newcommand{\ie}{\emph{i.e.}}
\newcommand{\eg}{\emph{e.g.}}

\setlist{nolistsep}

\algrenewcommand\algorithmicrequire{\textbf{Input:}}
\algrenewcommand\algorithmicensure{\textbf{Output:}}

\algnewcommand\Input{\item[\algorithmicinput]}
\algnewcommand\Output{\it em[\algorithmicoutput]}

\newcommand{\deltashap}{DeltaSHAP}
\newcommand{\ceil}[1]{\left\lceil #1 \right\rceil}

\definecolor{lightgray}{gray}{0.9}
\definecolor{gray}{gray}{0.7}
\definecolor{darkgray}{gray}{0.5}
\definecolor{custompink}{rgb}{0.99,0.03,0.51}

\icmltitlerunning{
DeltaSHAP: Explaining Prediction Evolutions in Online Patient Monitoring with Shapley Values
}

\begin{document}

\twocolumn[
\icmltitle{
\deltashap: Explaining Prediction Evolutions \\ in Online Patient Monitoring with Shapley Values
}

\icmlsetsymbol{equal}{*}

\begin{icmlauthorlist}
\icmlauthor{Changhun Kim}{equal,aitrics,kaist}
\icmlauthor{Yechan Mun}{equal,aitrics}
\icmlauthor{Sangchul Hahn}{aitrics}
\icmlauthor{Eunho Yang}{aitrics,kaist}
\end{icmlauthorlist}

\icmlaffiliation{aitrics}{AITRICS, Seoul, South Korea}
\icmlaffiliation{kaist}{Kim Jaechul Graduate School of AI, KAIST, Daejeon, South Korea}
\icmlcorrespondingauthor{Eunho Yang}{eunhoy@aitrics.com}

\icmlkeywords{Online Patient Monitoring, Time Series, XAI, Explainability}
\vskip 0.3in
]

\printAffiliationsAndNotice{\icmlEqualContribution}

\begin{abstract}
This study proposes \deltashap{}, a novel explainable artificial intelligence (XAI) algorithm specifically designed for online patient monitoring systems. In clinical environments, discovering the causes driving patient risk evolution is critical for timely intervention, yet existing XAI methods fail to address the unique requirements of clinical time series explanation tasks.
To this end, \deltashap{} addresses three key clinical needs: explaining the changes in the consecutive predictions rather than isolated prediction scores, providing both magnitude and direction of feature attributions, and delivering these insights in real time. By adapting Shapley values to temporal settings, our approach accurately captures feature coalition effects. It further attributes prediction changes using only the actually observed feature combinations, making it efficient and practical for time-sensitive clinical applications.
We also introduce new evaluation metrics to evaluate the faithfulness of the attributions for online time series, and demonstrate through experiments on online patient monitoring tasks that \deltashap{} outperforms state-of-the-art XAI methods in both explanation quality as 62\% and computational efficiency as 33\% time reduction on the MIMIC-III decompensation benchmark. We release our code at \url{https://github.com/AITRICS/DeltaSHAP}.
\end{abstract}
\section{Introduction}\label{sec:introduction}

Identifying patients at high risk of deterioration before critical events occur is a pivotal undertaking for clinical decision support systems~\citep{rothman2013development,churpek2016multicenter}. These early prediction systems often necessitate high transparency, as clinicians must understand the rationale behind alerts to trust and act upon them appropriately. Although deep neural networks have demonstrated remarkable performance and have become the \emph{de facto} standard within the healthcare domain~\citep{lstm,rajkomar2018scalable,che2018recurrent,seft,strats,mtand}, they often face resistance in clinical environments where accountability and explainability are paramount, primarily due to their black-box nature~\citep{ahmad2018interpretable}. Hence, establishing adequate explainable artificial intelligence (XAI) systems that address the unique characteristics of clinical environments remains an urgent and significant challenge.

Developing XAI for online patient monitoring presents additional challenges beyond those in static prediction settings such as image classification or tabular risk scoring. These challenges arise from the nature of clinical time series data—continuous monitoring, temporal dependencies, and multivariate signals such as vital signs and lab results. XAI methods must therefore be tailored to reflect the dynamic, sequential nature of patient care.
First, \emph{explaining prediction differences between consecutive time steps is essential}, as practitioners mainly focus on these temporal changes rather than absolute predictions at isolated time points; this is attributed to the fact that identical absolute predictions may indicate vastly different contextual meanings. For example, a decrease in sepsis probability from 70\% to 40\% indicates patient improvement, while an increase from 10\% to 40\% signals severe deterioration.
Second, in these situations, \emph{clinicians require both the magnitude and direction of feature attributions based on recent changes in patient measurements}, as such changes can either increase or decrease the prediction score. This directional information is clinically relevant, as individual features may indicate either deterioration or stabilization of the patient's condition.
Third, \emph{these attributions must be calculated in real-time} to enable rapid assessment and intervention. From these perspectives, na\"ively adopting conventional modality-agnostic XAI methods~\citep{lime,shap,ig,deeplift,fo} fall short in all these aspects as they focus solely on estimating attributions for absolute predictions while attempting to calculate pointwise attributions across all features and time steps, which is computationally burdensome.

Although not specifically designed for clinical monitoring, recent advances in time series XAI methods---FIT~\citep{fit} and WinIT~\citep{winit}---have emerged to address online prediction tasks.
FIT~\citep{fit} quantifies feature attributions based on their contribution to the prediction differences between the current time point and the preceding measurement. WinIT~\citep{winit} extends this approach by estimating temporal dependencies by aggregating its delayed impact across multiple consecutive time steps within fixed prediction windows.
While these methods decently identify important features, they are still commonly limited in that they can provide only the \emph{magnitude} of attributions while neglecting the \emph{directional effects} of the features for the prediction---precisely the information clinicians need most for decision-making. Also, their reliance on counterfactual generation strategies utilizing condition generative models significantly increases computation costs, making them impractical in time-sensitive clinical settings.

Motivated by these limitations, we propose \deltashap, a new XAI algorithm specifically designed to explain prediction differences between consecutive time steps in online patient monitoring, satisfying \emph{all critical requirements above}. \deltashap{} compares scenarios in which features at the most recent time step are either unobserved or fully observed, estimating directional attributions solely for these newly observed features and expressing the resulting prediction change as a sum of individual contributions. It efficiently approximates each feature’s contribution via Shapley value sampling, capturing feature interactions without the exponential cost. To improve efficiency in clinical settings, \deltashap{} evaluates only observed feature subsets, naturally handling irregular data and avoiding unrealistic imputations. Moreover, it further reduces overhead by leveraging simple missing-value strategies, such as forward-filling, instead of relying on costly generative models. As a fully model-agnostic method, \deltashap{} requires no access to gradients or internal operations, enabling broad compatibility with diverse architectures and preprocessing pipelines.

We further introduce novel evaluation metrics that quantify attribution faithfulness in online time series prediction by measuring the area under curves as the most and least important features are progressively removed. These metrics—capturing prediction differences, AUC drop, and APR drop---enable comprehensive benchmarking. Extensive experiments on MIMIC-III decompensation and PhysioNet 2019 sepsis prediction tasks~\citep{mimic3,reyna2020early}, using a representative LSTM backbone~\citep{lstm}, demonstrate that \deltashap{} identifies important features more effectively and efficiently, achieving up to a 62\% improvement while reducing runtime by 33\%.

To summarize, our contribution is threefold:

\begin{itemize}
    \item We propose \deltashap, a novel XAI method for online patient monitoring that explains prediction differences between consecutive time steps, providing directional attributions efficiently.

    \item We further propose novel evaluation metrics to assess attribution faithfulness in online monitoring scenarios, addressing the lack of standardized benchmarks for dynamic prediction settings.

    \item Through extensive experiments on real-world clinical datasets, we demonstrate that \deltashap{} outperforms existing XAI methods in both attribution accuracy and computational efficiency.

\end{itemize}
\section{Related Work} \label{sec:related_work}
\paragraph{Modality-agnostic XAI.}
Despite achieving remarkable performance across diverse domains, deep neural networks largely operate as opaque black boxes that reveal little about their internal reasoning processes~\citep{christoph2020interpretable}. This lack of transparency undermines trust and hampers accountability, particularly in safety-critical domains like healthcare, where decisions can lead to far-reaching consequences.
In response, various modality-agnostic explainable artificial intelligence (XAI) approaches have been developed. Attribution-based methods such as LIME~\citep{lime} and SHAP~\citep{shapley,shap}, as well as its variants like KernalSHAP, GradientSHAP, and DeepSHAP~\citep{shap,captum} map model predictions to specific input features, quantifying both the strength and direction of each feature’s contribution to the final output.
Meanwhile, gradient-based methods such as Integrated Gradients (IG)~\citep{ig} and DeepLIFT~\citep{deeplift} estimate feature attributions by analyzing the gradients of the model with respect to its inputs. Complementary to these, perturbation-based approaches provide alternative insights into feature importance.
Feature Occlusion (FO)~\citep{fo} assesses importance by systematically replacing individual features with zeros or random noise and observing prediction changes. Augmented Feature Occlusion (AFO)~\citep{fit} refines this method by replacing features with values sampled from the training dataset, which helps maintain data distribution properties.
While these XAI methods have improved model transparency, they have been primarily validated in computer vision tasks~\citep{das2020opportunities}. However, their application to online patient monitoring remains underexplored---a notable limitation, as such modality-agnostic XAI approaches often overlook the temporal dependencies inherent in sequential clinical data.

\paragraph{XAI for online time series.}
To address the complex challenges of developing XAI for temporal data, researchers have proposed numerous specialized time series XAI algorithms~\citep{timeshap,fit,winit,dynamask,extrmask,contralsp,timex,timex++}. Among these approaches, FIT~\citep{fit} and WinIT~\citep{winit} stand out for their specific focus on attribution estimation in online time series prediction---the primary concern of our research.
FIT~\citep{fit} quantifies feature attributions by analyzing how observations contribute to prediction changes between sequential time points. In doing so, it employs a KL-divergence framework that compares the predictive distribution against counterfactuals where specific features remain unobserved, enabling the detection of subtle shifts in patient status. Building on this foundation, WinIT~\citep{winit} extends the approach by capturing each observation's delayed impacts across multiple future time steps, recognizing that clinical measurements frequently influence predictions beyond immediate timeframes. It further explicitly models temporal dependencies by calculating importance through differential scoring that accounts for sequential observation relationships.
Despite their advances in identifying relevant features, they share significant limitations for clinical applications. They primarily provide attribution magnitude without indicating \emph{directional effects}---whether an increasing heart rate positively or negatively influences the prediction outcome, which is essential information for clinical decision-making. Furthermore, their dependence on counterfactual generation through conditional generative models introduces \emph{substantial computational overhead}, rendering them impractical for time-sensitive clinical environments where rapid explanation delivery is crucial for timely intervention. These shortcomings motivate our development of \deltashap, which fundamentally reimagines temporal attribution by providing \emph{directional feature importance} while maintaining \emph{computational efficiency}---critical requirements for effective clinical decision support systems.
\section{Preliminaries} \label{sec:preliminaries}
This section sets the stage for \deltashap{} by outlining the necessary preliminaries. We begin with the problem formulation of estimating attributions at the most recent time step in~\Cref{subsec:problem_setup}, followed by a review of classical Shapley values in~\Cref{subsec:shapley_values}, both of which serve as the methodological foundation of our method.

\subsection{Problem Setup} \label{subsec:problem_setup}
Let $f: \mathbb{R}^{L \times D} \rightarrow [0, 1]$ be an online patient monitoring model, where $L$ represents the maximum sequence length, $D$ is the number of clinical features, and the output is the probability of a critical outcome within a clinically actionable timeframe (\eg, 24 or 48 hours). At each time step $T$, the model processes patient data of vital signs and laboratory values using a sliding window approach, taking the most recent window $\mathbf{X}_{T-W+1:T} \in \mathbb{R}^{W \times D}$ from the complete record $\mathbf{X} \in \mathbb{R}^{L \times D}$.

We denote $f(\mathbf{X}_{T-W+1:T} \setminus \mathbf{X}_T)$ as the prediction using only historical data without current measurements. Then, our objective is to explain the prediction difference:
\begin{equation*}
    \Delta = f(\mathbf{X}_{T-W+1:T}) - f(\mathbf{X}_{T-W+1:T} \setminus \mathbf{X}_T).
\end{equation*}
This requires computing attributions $\phi(f, \mathbf{X}_{T-W+1:T}) \in \mathbb{R}^{D}$ for features at the current time step, where $\phi_j(f, \mathbf{X}_{T-W+1:T})$ quantifies how feature $j$ at time $T$ contributes to the prediction evolution of $\Delta$. These attributions are \emph{directional}, with positive values indicating features that increase the prediction and negative values showing features that decrease it, while satisfying the fundamental efficiency property:
\begin{equation*} \label{eqn:deltashapproperty}
    \sum_{j=1}^{D}\phi_j(f, \mathbf{X}_{T - W + 1:T}) = \Delta.
\end{equation*}

\begin{figure*}[!t]
\centering
\includegraphics[width=\textwidth]{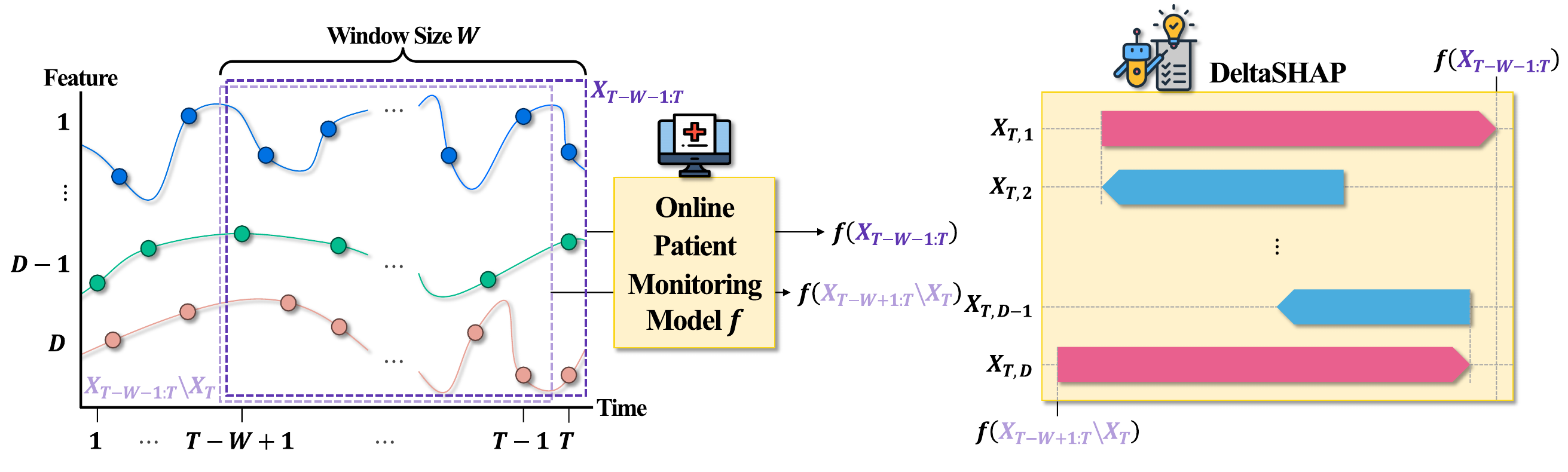}
\vspace{-.3in}
\caption{
    Overview of the \deltashap\ framework for explaining prediction differences in online patient monitoring. Unlike existing methods, \deltashap\ delivers both the magnitude and directional effects of feature attributions by comparing scenarios where recent features are either unobserved or fully observed. By adapting Shapley values to temporal settings, it efficiently processes irregular measurements without requiring expensive generative models, enabling real-time explanations critical for clinical decision-making.
}
\label{fig:main_figure}
\vspace{-.2in}
\end{figure*}

\subsection{Shapley Values} \label{subsec:shapley_values}
Originally developed in cooperative game theory, Shapley values~\citep{shapley} provide a principled framework for fairly allocating contributions among participants. In the context of XAI for time series, they enable an effective quantification of each feature’s influence at the current time step $T$, while naturally accounting for complex dependencies and interactions among features.
Given our focus on explaining $\Delta$, we define Shapley values for feature $j$ at time $T$ as:
\begin{equation} \label{eqn:shap}
\phi_j(f, \mathbf{X}) = \frac{1}{D} \sum_{S \subset \mathcal{F} \setminus \{j\}} 
\binom{D-1}{|S|}^{-1} \left[ v(S \cup \{j\}) - v(S) \right].
\end{equation}

Here, $\mathcal{F} = \{1, \dots, D\}$ represents the set of all features at time $T$, $S$ is a subset of these features, and $v(S)$ measures the partial contribution to $\Delta$ when only features in $S$ are observed at time $T$. The combinatorial weight $\binom{D-1}{|S|}^{-1}$ represents the fraction of all feature orderings in which subset $S$ precedes feature $j$, ensuring that the marginal contribution of $j$ is fairly averaged across all possible arrival orders.
If we define $\mathbf{X}_T^S$ as the observation at time $T$ where only features in subset $S$ are present (with others unobserved), then:
\begin{equation*}
    v(S) = f(\mathbf{X}_{T-W+1:T-1} \cup \mathbf{X}_T^S) - f(\mathbf{X}_{T-W+1:T} \setminus \mathbf{X}_T).
\end{equation*}

\section{DeltaSHAP} \label{sec:method}
This section presents \deltashap{}, a new XAI framework designed to explain prediction evolutions in online patient monitoring.
An overview of the \deltashap{} framework is shown in~\Cref{fig:main_figure}, with the full algorithm detailed in~\Cref{alg:deltashap}.

Unlike conventional SHAP, which computes attributions across all time steps using all-zero baselines, \deltashap{} focuses on prediction differences between consecutive steps by comparing cases where newly observed features at the final time step are either missing or present, expressing this as a directional sum of their attributions~(\Cref{subsec:deltashap}).
To reduce the computational burden of evaluating feature subsets, \deltashap{} adopts Shapley value sampling through feature permutations~\citep{mitchell2022sampling,strumbelj2010efficient}~(\Cref{subsec:shapley_value_sampling}).
Instead of relying on complex generative models, it leverages simple missing-value handling strategies, such as forward-filling for LSTM~\citep{lstm}, resulting in substantial computational savings~(\Cref{subsec:baseline_selection}).

\subsection{Explaining Prediction Evolutions} \label{subsec:deltashap}
As illustrated in~\Cref{sec:preliminaries}, our goal is to explain the prediction difference $\Delta = f(\mathbf{X}_{T-W+1:T}) - f(\mathbf{X}_{T-W+1:T} \setminus \mathbf{X}_T)$ between two consecutive time steps $T - 1$ and $T$ by estimating directional attributions $\phi(f, \mathbf{X}_{T-W+1:T}) \in \mathbb{R}^{D}$ for features at the current time step $T$. We denote the attribution for the $j$-th feature as $\phi_j(f, \mathbf{X}_{T-W+1:T})$, with the vector
\begin{equation*}
    \phi(f, \mathbf{X}_{T-W+1:T}) = \left( \phi_j(f, \mathbf{X}_{T-W+1:T}) \right)_{j=1}^{D}.
\end{equation*}
These attributions satisfy the fundamental efficiency property $\sum_{j=1}^{D}\phi_j(f, \mathbf{X}_{T - W + 1}) = \Delta$ as in~\Cref{eqn:deltashapproperty}.

However, the exponential complexity of the exact Shapley value computation, as defined in~\Cref{eqn:shap}, renders exhaustive evaluation impractical for high-dimensional clinical data. Consequently, we exploit a Shapley Value Sampling approximation technique~\citep{mitchell2022sampling,strumbelj2010efficient} to efficiently estimate these attributions while maintaining the core mathematical principles of feature contribution, as detailed in~\Cref{subsec:shapley_value_sampling}.

\subsection{Shapley Value Sampling} \label{subsec:shapley_value_sampling}
Calculating exact Shapley values requires evaluating $2^D$ possible feature combinations within the last time step, which incurs substantial computational cost when applied to high-dimensional clinical data. We address this challenge by employing a Shapley Value Sampling approach that randomly permutes features, considering the irregular observation of clinical data.
Specifically, we sample $N$ random permutations $\Omega$ of the total feature set, approximating Shapley values for each observed feature $j \in \mathcal{F}_{\text{obs}}$:
\begin{equation*} \label{eqn:deltashap}
    \hat{\phi}_j(f, \mathbf{X}_{T-W+1:T}) = \frac{1}{N} \sum_{\pi \in \Omega} [v(S_{\pi,j} \cup \{j\}) - v(S_{\pi,j})],
\end{equation*}
where $S_{\pi, j} \subset \mathcal{F}_{\text{obs}}$ denotes the subset of features that precede feature $j$ in permutation $\pi$, and $v(S_{\pi, j})$ evaluates the model’s output difference $\Delta$ when only the features in $S_{\pi, j}$ are considered observed at the final time step. This formulation allows us to efficiently estimate the marginal contribution of each feature by averaging over its contextual impact across different arrival orders, consistent with Shapley value theory. To mitigate sampling error and ensure the efficiency property, we normalize the attributions:
\begin{equation} \label{eqn:normalization}
\begin{split}
    & \phi_j(f, \mathbf{X}_{T-W+1:T}) \\
    & = \hat{\phi}_j(f, \mathbf{X}_{T-W+1:T}) \cdot \frac{\Delta}{\sum_{k \in \mathcal{F}_{\text{obs}}} \hat{\phi}_k(f, \mathbf{X}_{T-W+1:T})},
\end{split}
\end{equation}
where $\Delta$ represents the total prediction difference, and the denominator ensures that the total attributed change exactly matches the observed prediction difference.

This sampling-based approach provides an \emph{unbiased estimate of exact Shapley values} while reducing the computational complexity from exponential to linear in the number of permutations $N$. The normalization step not only mitigates sampling variability but also enforces the efficiency property, ensuring that the total attribution matches the prediction difference without altering the relative importance or ranking of salient features---making \deltashap{} both principled and scalable for real-time clinical interpretation.

\subsection{Baseline Selection} \label{subsec:baseline_selection}
One natural question that arises in~\Cref{eqn:deltashap} is how to handle unobserved features. Existing approaches often use simple zero substitution~\citep{lime,shap,ig,deeplift,fo}, which fails to account for temporal dependencies in time series data, or employ conditional generative models~\citep{fit,winit}, which significantly increase computational costs while introducing imputation errors. Instead, we propose a simple yet effective baseline selection strategy that leverages existing missing data handling mechanisms from preprocessing. Specifically, for LSTM models~\citep{lstm}, we utilize \emph{forward-filling}, where missing values are imputed using the most recent observed value. This approach offers significant advantages by aligning with pre-processing methods, reducing out-of-distribution problems, and maintaining computational efficiency. By leveraging existing preprocessing techniques, we enable model-specific missing value handling while preserving a consistent attribution framework across different architectures.

\subsection{Practical Benefits of DeltaSHAP} \label{subsec:practical_benefits}
\deltashap{} offers several distinctive advantages for online patient monitoring.
First, \deltashap{} precisely captures clinical reasoning by representing prediction differences through a principled sum of signed feature attributions. By adapting Shapley values, it meticulously accounts for complex feature interactions, providing nuanced insights into how individual features contribute to clinical predictions.
Second, \deltashap{} offers significantly improved computational efficiency over the original SHAP~\citep{shap}. By strategically sampling Shapley values on the most recent features and using a simplified baseline, it avoids the overhead of generative-model-based alternatives~\citep{fit,winit}, enabling near-instantaneous explanations essential in time-sensitive clinical settings.
Last, the method's model-agnostic nature emerges from its fundamental design---operating purely on input-output relationships without penetrating model-specific internal representations. This versatility ensures \deltashap{} can seamlessly interpret predictions across diverse model architectures, making it an adaptable tool for comprehensive clinical decision support.
\section{Experiments} \label{sec:experiments}
This section evaluates \deltashap's performance against existing baselines and demonstrates its practical advantages. We elaborate our experimental setup in~\Cref{subsec:exp_setup}, followed by addressing three key research questions:
\begin{itemize}
    \item Does \deltashap{} outperform existing methods in attribution accuracy and computational efficiency?~(\Cref{subsec:main_results,subsec:efficiency})
    
    \item Do the individual components of \deltashap{} truly contribute to its overall performance gains? (\Cref{subsec:ablation_study})
    
    \item Are \deltashap's attributions coherent with clinical domain knowledge, providing interpretations that clinicians can trust? (\Cref{subsec:case_study})

\end{itemize}

\subsection{Experimental Setup} \label{subsec:exp_setup}
\paragraph{Datasets.}
We evaluate XAI methods for online patient monitoring using two datasets: MIMIC-III~\citep{mimic3} and PhysioNet 2019~\citep{reyna2020early}. Both datasets reflect real-world online monitoring scenarios, where multiple risk scores are generated per patient throughout hospitalization. The MIMIC-III decompensation prediction task~\citep{mimic3} includes ICU admissions from Beth Israel Deaconess Medical Center between 2001 and 2012, with the goal of predicting decompensation within 24 hours. From approximately 41,000 ICU stays, we generate 2.5 million prediction instances, of which 2.5\% (63,000) are labeled positive.
The PhysioNet 2019 early sepsis prediction challenge~\citep{reyna2020early} targets the detection of sepsis onset in ICU patients, following the Sepsis-3 criteria~\citep{singer2016third}, with a prediction window of 12 hours. From around 40,000 ICU admissions, we derive approximately 1.1 million prediction instances, with 2.5\% (27,000) labeled positive for sepsis.

\paragraph{Backbone architectures.}
To evaluate \deltashap{} and other XAI counterparts, we employ LSTM~\citep{lstm}, a widely used time series classification architecture. As a specialized RNN, LSTMs excel at modeling complex temporal patterns in clinical data by maintaining information over extended periods and selectively updating internal states, making them fundamental in numerous patient monitoring systems where physiological signal dynamics are crucial. Demonstrating explainability in LSTM provides a solid foundation for evaluating the reliability of XAI methods in healthcare time series applications.

\paragraph{XAI baselines.}
We utilize the following baseline methods as comparison models for our approach. We first implement modality-agnostic XAI methods that explain model behavior by systematically perturbing inputs, including LIME~\citep{lime}, GradSHAP~\citep{shap}, which is a representative method within the SHAP variants, FO~\citep{fo}, and AFO~\citep{fit}. We also compare with modality-agnostic approaches leveraging model gradients to determine feature importance, represented by IG~\citep{ig} and DeepLIFT~\citep{deeplift}. Last, we include specialized techniques designed for explaining time series predictions in real-time applications, including FIT~\citep{fit} and WinIT~\citep{winit}.

\paragraph{Evaluation metrics.}

Although several XAI methods have been proposed for online time series prediction~\citep{fit,winit}, they are typically evaluated in offline settings with a single prediction per time series---most notably on the MIMIC-III mortality task~\citep{mimic3}. However, no standard exists for evaluating XAI in online patient monitoring. To address this, we propose new metrics, starting with Cumulative Prediction Difference (CPD) and Cumulative Prediction Preservation (CPP) from~\citet{jang2025timing}. For a model $f$ and input $X_{T-W+1:T}$:
\begin{align*}
\text{CPD}(f, X, K) &= \sum_{k=0}^{K-1} \left|f(X^{\uparrow}_k) - f(X^{\uparrow}_{k+1}) \right|, \\
\text{CPP}(f, X, K) &= \sum_{k=0}^{K-1} \left|f(X^{\downarrow}_k) - f(X^{\downarrow}_{k+1}) \right|,
\end{align*}
where $X^{\uparrow}_k$ and $X^{\downarrow}_k$ are derived by removing the top-$k$ and bottom-$k$ features (by absolute attribution) at the final time step $T$. This stepwise removal avoids the \emph{cancel-out problem}, where opposing attributions cancel each other out and distort faithfulness metrics. In summary, these metrics capture marginal effects more reliably and improve the evaluation of directional attributions in online settings.

Based on these foundations, we propose our area-based primary metrics of Area Under Prediction Difference (AUPD) and Area Under Prediction Preservation (AUPP):
\begin{align*}
    \text{AUPD}(f, X, K) &= \frac{1}{K} \sum_{k=1}^{K} \text{CPD}(f, X, k), \\
    \text{AUPP}(f, X, K) &= \frac{1}{K} \sum_{k=1}^{K} \text{CPP}(f, X, k).
\end{align*}
These metrics offer two main advantages over CPD and CPP. First, they assess attribution ordering better by emphasizing the influence of higher-ranked features. Second, they reduce sensitivity to local anomalies by aggregating effects across all top-$k$ (or bottom-$k$) removals. As a result, AUPD and AUPP yield more stable and faithful evaluations of attribution quality in dynamic online settings.

\begin{table*}[!t]
\centering
\caption{
Performance comparison of XAI methods on clinical prediction tasks: MIMIC-III decompensation (top) and PhysioNet 2019 sepsis (bottom) benchmarks using LSTM as backbone architecture. We evaluate by removing the most or least salient 25\% of the features per time step with forward fill substitution.
}
\label{tab:main_lstm}
\vspace{-.1in}
\resizebox{\textwidth}{!}{
\setlength{\tabcolsep}{5pt}
\begin{tabular}{l|ccc|ccc|c}
    \toprule

    \multirow{2}{*}{\textbf{Algorithm}} & \multicolumn{3}{c|}{\textbf{Removal of Most Salient 25\% Features}} & \multicolumn{3}{c|}{\textbf{Removal of Least Salient 25\% Features}} & \multirow{2}{*}{\textbf{Wall-Clock Time}} \\
    
    & AUPD $\uparrow$ & AUAUCD $\uparrow$ & AUAPRD $\uparrow$ & AUPP $\downarrow$ & AUAUCP $\downarrow$ & AUAPRP $\downarrow$ & \\
    
    \midrule
    
    LIME~\citep{lime}      & 8.20\scriptsize{$\pm$0.03} & 274.03\scriptsize{$\pm$4.95} & 1012.26\scriptsize{$\pm$22.89} & 21.58\scriptsize{$\pm$0.03} & 523.93\scriptsize{$\pm$4.71} & 1136.19\scriptsize{$\pm$17.84} & 0.22 \\
    
    GradSHAP~\citep{shap} & 6.20\scriptsize{$\pm$0.02} & 120.45\scriptsize{$\pm$5.08} & 671.54\scriptsize{$\pm$7.87} & 19.68\scriptsize{$\pm$0.03} & 281.43\scriptsize{$\pm$2.84} & 1209.98\scriptsize{$\pm$10.28} & \underline{0.03} \\
    
    IG~\citep{ig}        & 13.46\scriptsize{$\pm$0.00} & \underline{404.24\scriptsize{$\pm$0.00}} & \underline{1408.36\scriptsize{$\pm$0.00}} & 14.51\scriptsize{$\pm$0.00} & 103.73\scriptsize{$\pm$0.00} & \underline{420.69\scriptsize{$\pm$0.00}} & 0.04 \\
    
    DeepLIFT~\citep{deeplift}  & \underline{13.95\scriptsize{$\pm$0.00}} & 402.70\scriptsize{$\pm$0.00} & 1376.30\scriptsize{$\pm$0.00} & 14.35\scriptsize{$\pm$0.00} & 109.61\scriptsize{$\pm$0.00} & 472.90\scriptsize{$\pm$0.00} & \underline{0.03} \\
    
    FO~\citep{fo}        & 13.55\scriptsize{$\pm$0.00} & 326.52\scriptsize{$\pm$0.00} & 1365.62\scriptsize{$\pm$0.00} & \underline{14.14\scriptsize{$\pm$0.00}} & 120.84\scriptsize{$\pm$0.00} & 447.89\scriptsize{$\pm$0.00} & 1.43 \\
    
    AFO~\citep{fit}       & 13.08\scriptsize{$\pm$0.05} & 311.16\scriptsize{$\pm$2.29} & 1283.98\scriptsize{$\pm$6.59} & 15.14\scriptsize{$\pm$0.04} & 139.12\scriptsize{$\pm$4.13} & 618.23\scriptsize{$\pm$21.72} & 39.62 \\
    
    FIT~\citep{fit}       & 12.60\scriptsize{$\pm$0.00} & \textbf{407.65\scriptsize{$\pm$0.00}} & 1301.38\scriptsize{$\pm$0.00} & 16.16\scriptsize{$\pm$0.00} & \underline{84.90\scriptsize{$\pm$0.00}} & 550.40\scriptsize{$\pm$0.00} & 0.12 \\
    
    WinIT~\citep{winit}     & 10.06\scriptsize{$\pm$1.48} & 202.09\scriptsize{$\pm$20.61} & 1012.61\scriptsize{$\pm$98.87} & 16.56\scriptsize{$\pm$1.75} & 234.39\scriptsize{$\pm$38.93} & 822.12\scriptsize{$\pm$146.30} & 0.30 \\
    
    \midrule
    
    \rowcolor{lightgray} \deltashap\ & \textbf{22.59\scriptsize{$\pm$0.01}} & 395.08\scriptsize{$\pm$1.98} & \textbf{1558.28\scriptsize{$\pm$8.61}} & \textbf{3.04\scriptsize{$\pm$0.01}} & \textbf{65.06\scriptsize{$\pm$1.04}} & \textbf{242.50\scriptsize{$\pm$7.41}} & \textbf{0.02} \\
    
    \midrule
    
    \midrule

    \multirow{2}{*}{\textbf{Algorithm}} & \multicolumn{3}{c|}{\textbf{Removal of Most Salient 25\% Features}} & \multicolumn{3}{c|}{\textbf{Removal of Least Salient 25\% Features}} & \multirow{2}{*}{\textbf{Wall-Clock Time}} \\
    
    & AUPD $\uparrow$ & AUAUCD $\uparrow$ & AUAPRD $\uparrow$ & AUPP $\downarrow$ & AUAUCP $\downarrow$ & AUAPRP $\downarrow$ & \\
    
    \midrule
    
    LIME~\citep{lime}      & 1.13\scriptsize{$\pm$0.00} & 132.77\scriptsize{$\pm$5.47} & 11.88\scriptsize{$\pm$0.11} & 3.58\scriptsize{$\pm$0.00} & 122.11\scriptsize{$\pm$1.34} & 15.52\scriptsize{$\pm$1.03} & 0.29 \\
    
    GradSHAP~\citep{shap} & 0.96\scriptsize{$\pm$0.00} & 103.91\scriptsize{$\pm$2.84} & 9.02\scriptsize{$\pm$0.55} & 3.09\scriptsize{$\pm$0.00} & 98.07\scriptsize{$\pm$6.36} & 13.55\scriptsize{$\pm$0.95} & 0.04 \\
    
    IG~\citep{ig}        & 2.28\scriptsize{$\pm$0.00} & 176.18\scriptsize{$\pm$0.00} & 18.84\scriptsize{$\pm$0.00} & 2.42\scriptsize{$\pm$0.00} & 47.21\scriptsize{$\pm$0.00} & 8.60\scriptsize{$\pm$0.00} & 0.04 \\
    
    DeepLIFT~\citep{deeplift}  & 2.24\scriptsize{$\pm$0.00} & 171.10\scriptsize{$\pm$0.00} & 17.04\scriptsize{$\pm$0.00} & 2.45\scriptsize{$\pm$0.00} & 65.75\scriptsize{$\pm$0.00} & 9.90\scriptsize{$\pm$0.00} & \underline{0.03} \\
    
    FO~\citep{fo} & 2.34\scriptsize{$\pm$0.00} & 195.34\scriptsize{$\pm$0.00} & 21.06\scriptsize{$\pm$0.00} & 2.37\scriptsize{$\pm$0.00} & 49.70\scriptsize{$\pm$0.00} & 9.02\scriptsize{$\pm$0.00} & 1.14 \\
    
    AFO~\citep{fit}      & \underline{3.27\scriptsize{$\pm$0.00}} & 207.51\scriptsize{$\pm$6.99} & \textbf{24.17\scriptsize{$\pm$0.26}} & \underline{1.03\scriptsize{$\pm$0.00}} & \underline{24.64\scriptsize{$\pm$7.70}} & \underline{1.43\scriptsize{$\pm$0.25}} & 14.18 \\
    
    FIT~\citep{fit}      & 2.15\scriptsize{$\pm$0.00} & \underline{208.82\scriptsize{$\pm$0.00}} & \underline{24.16\scriptsize{$\pm$0.00}} & 3.08\scriptsize{$\pm$0.00} & 58.77\scriptsize{$\pm$0.00} & 4.70\scriptsize{$\pm$0.00} & 0.11 \\
    
    WinIT~\citep{winit}     & 1.27\scriptsize{$\pm$0.00} & 135.51\scriptsize{$\pm$0.28} & 13.75\scriptsize{$\pm$0.10} & 3.23\scriptsize{$\pm$0.00} & 96.50\scriptsize{$\pm$0.66} & 14.50\scriptsize{$\pm$0.21} & 0.29 \\
    
    \midrule
    
    \rowcolor{lightgray} \deltashap\ & \textbf{3.68\scriptsize{$\pm$0.00}} & \textbf{219.11\scriptsize{$\pm$1.75}} & 22.49\scriptsize{$\pm$0.13} & \textbf{0.89\scriptsize{$\pm$0.00}} & \textbf{8.46\scriptsize{$\pm$1.95}} & \textbf{0.25\scriptsize{$\pm$0.08}} & \textbf{0.02} \\
    
    \bottomrule
\end{tabular}}
\end{table*}

Next, considering that the area under the receiver operating characteristic curve (AUC) and the area under the precision-recall curve (APR) are frequently used in online patient monitoring model performance reporting~\citep{fit,winit,seft}, we further extend these concepts to develop additional metrics that incorporate performance measurements across the entire test set $\mathcal{D}_{\text{test}}$:
\begin{align*}
&\begin{split}
    &\text{AUAUCD}(f, \mathcal{D}_{\text{test}}, K) \\
    &= \frac{1}{K} \sum_{k=1}^{K} \left|\text{AUC}(f, \mathcal{D}_{\text{test}}) - \text{AUC}(f, \mathcal{D}_{\text{test}, k}^{\uparrow}) \right|,
\end{split} \\
&\begin{split}
    &\text{AUAPRD}(f, \mathcal{D}_{\text{test}}, K) \\
    &= \frac{1}{K} \sum_{k=1}^{K} \left|\text{APR}(f, \mathcal{D}_{\text{test}}) - \text{APR}(f, \mathcal{D}_{\text{test}, k}^{\uparrow}) \right|,
\end{split} \\
&\begin{split}
    &\text{AUAUCP}(f, \mathcal{D}_{\text{test}}, K) \\
    &= \frac{1}{K} \sum_{k=1}^{K} \left| \text{AUC}(f, \mathcal{D}_{\text{test}}) - \text{AUC}(f, \mathcal{D}_{\text{test}, k}^{\downarrow}) \right|,
\end{split} \\
&\begin{split}
    &\text{AUAPRP}(f, \mathcal{D}_{\text{test}}, K) \\
    &= \frac{1}{K} \sum_{k=1}^{K} \left| \text{APR}(f, \mathcal{D}_{\text{test}}) - \text{APR}(f, \mathcal{D}_{\text{test}, k}^{\downarrow}) \right|,
\end{split}
\end{align*}
where $\mathcal{D}_{\text{test}, k}^{\uparrow}$ represents the test dataset with the top-$k$ features removed from the last time step of each sample based on absolute attribution scores, and $\mathcal{D}_{\text{test}, k}^{\downarrow}$ represents the test dataset with the bottom-$k$ features removed. $\text{AUC}(f, \mathcal{D}_{\text{test}})$ calculates the AUC metric using model $f$ on all samples in $\mathcal{D}_{\text{test}}$. 

Since the number of observed features varies across observations when removing $K$ features from the last timestep, we adaptively remove at most $K = \ceil{p \times \mathcal{F}_{obs}}$ features to remove the most salient features, whereas removing at most
\begin{equation*}
K = \underbrace{D - \mathcal{F}_{\text{obs}}}_{\text{missing features}} 
    + \underbrace{\left\lceil p \cdot \mathcal{F}_{\text{obs}} \right\rceil}_{\text{least salient observed features}}
\end{equation*}
features for removing the least salient features.

These area-based performance metrics offer unique advantages: (1) they evaluate attribution impact at the dataset level, capturing how feature removals affect overall model performance, rather than individual prediction shifts; and (2) by operating in the decision space of AUC and APR, they directly reflect changes relevant to clinical risk stratification, offering more actionable insight than logit-based metrics.

\paragraph{Implementation details.}
Our \deltashap{} implementation includes a single hyperparameter, $N$, denoting the number of random permutations, which we set to 25 for all experiments. For evaluation metrics, we consistently use an adaptive feature removal threshold of $p = 0.25$. In all tables, results are scaled by $10^4$ for readability, except for wall-clock time (reported in seconds per batch). The results are reported as mean~$\pm$~standard error over five repetitions. \textbf{Bold} and \underline{underlined} entries indicate the best and second-best performing methods, respectively. The complete implementation, including code to reproduce all experiments, is publicly available at \url{https://github.com/AITRICS/DeltaSHAP}.

\begin{table*}[!t]
\centering
\caption{
Ablation study of \deltashap\ with respect to its Baseline Selection (BS), Normalization (Norm.), and the number of permutations $N$ on MIMIC-III decompensation~\citep{mimic3} benchmark using LSTM~\citep{lstm} as backbone architecture.
We evaluate by removing the most or least salient 25\% of the features per time step with forward fill substitution.
}
\vspace{-.1in}
\label{tab:ablation_study}
\resizebox{\textwidth}{!}{
\setlength{\tabcolsep}{5pt}
\begin{tabular}{l|ccc|ccc|c}
    \toprule
    \multirow{2}{*}{\textbf{Algorithm}} & \multicolumn{3}{c|}{\textbf{Removal of Most Salient 25\% Features}} & \multicolumn{3}{c|}{\textbf{Removal of Least Salient 25\% Features}} & \multirow{2}{*}{\textbf{Wall-Clock Time}} \\
    
    & AUPD $\uparrow$ & AUAUCD $\uparrow$ & AUAPRD $\uparrow$ & AUPP $\downarrow$ & AUAUCP $\downarrow$ & AUAPRP $\downarrow$ & \\
    
    \midrule
    
    w/o BS & 8.49\scriptsize{$\pm$0.02} & 222.53\scriptsize{$\pm$4.17} & 957.08\scriptsize{$\pm$7.15} & 18.89\scriptsize{$\pm$0.02} & 234.09\scriptsize{$\pm$3.13} & 889.89\scriptsize{$\pm$7.83} & 0.05 \\
    
    w/o Norm. & \underline{22.58\scriptsize{$\pm$0.01}} & \underline{397.88\scriptsize{$\pm$1.96}} & \underline{1558.56\scriptsize{$\pm$8.60}} & \underline{3.05\scriptsize{$\pm$0.01}} & \textbf{60.57\scriptsize{$\pm$2.11}} & \underline{250.87\scriptsize{$\pm$10.04}} & 0.05 \\

    $N = 1$ & 22.14\scriptsize{$\pm$0.03} & 376.54\scriptsize{$\pm$6.34} & 1536.39\scriptsize{$\pm$9.30} & 3.19\scriptsize{$\pm$0.01} & 71.90\scriptsize{$\pm$3.86} & 264.40\scriptsize{$\pm$11.23} & \textbf{0.02} \\

    $N = 10$ & 22.56\scriptsize{$\pm$0.01} & 395.12\scriptsize{$\pm$3.45} & 1558.09\scriptsize{$\pm$10.04} & 3.07\scriptsize{$\pm$0.01} & \underline{61.35\scriptsize{$\pm$2.05}} & 253.15\scriptsize{$\pm$7.69} & \underline{0.04} \\

    $N = 100$ & \textbf{22.61\scriptsize{$\pm$0.00}} & \textbf{398.62\scriptsize{$\pm$1.38}} & \textbf{1561.30\scriptsize{$\pm$6.44}} & \textbf{3.04\scriptsize{$\pm$0.00}} & 64.77\scriptsize{$\pm$1.30} & \textbf{246.42\scriptsize{$\pm$0.63}} & 0.09 \\
    
    \midrule
    
    \rowcolor{lightgray} \deltashap\ & \underline{22.58\scriptsize{$\pm$0.01}} & \underline{397.88\scriptsize{$\pm$1.96}} & \underline{1558.56\scriptsize{$\pm$8.60}} & \underline{3.05\scriptsize{$\pm$0.01}} & \textbf{60.57\scriptsize{$\pm$2.11}} & \underline{250.87\scriptsize{$\pm$10.04}} & 0.05 \\
    
    \bottomrule
\end{tabular}}
\end{table*}

\subsection{Main Results} \label{subsec:main_results}
This subsection evaluates \deltashap{}'s faithfulness compared to existing attribution methods. Results in~\Cref{tab:main_lstm} demonstrate that \deltashap{} consistently outperforms all baselines across both datasets---MIMIC-III~\citep{mimic3} and PhysioNet 2019~\citep{reyna2020early}. \deltashap{} achieves superior scores in AUPD and AUPP metrics, which directly measure explanation alignment with model behavior rather than ground truth dependency. Particularly noteworthy is \deltashap{}'s substantial performance margin in AUPD, AUPP, AUAUCP, and AUAPRP---likely resulting from Shapley values' effective capture of feature interactions, focused last-timestep attribution, and preprocessing-aligned baseline selection.
It is also worth highlighting that \deltashap{} is the only method where AUPD changes exceed AUPP changes across all model-dataset combinations, underscoring its exceptional capacity for model-aligned explanations and demonstrating its superiority.

\subsection{Computational Efficiency Analysis} \label{subsec:efficiency}
In this subsection, we evaluate the computational efficiency of \deltashap{} against existing baselines. To this end, we measure wall-clock time (seconds per batch) on the MIMIC-III decompensation benchmark using LSTM on a single GPU with batch size 1. \Cref{tab:main_lstm} showcases that \deltashap{} achieves superior execution time, about 0.02 seconds/batch, through parallel forwarding across all feature occlusion samples and permutations. In contrast, competing methods require either computationally expensive multiple forward passes~\citep{lime,shap,fo,fit,winit} or gradient calculations via backward passes~\citep{ig,deeplift}. This efficiency is particularly valuable for clinical online patient monitoring, where rapid intervention is critical.

\subsection{Ablation Study} \label{subsec:ablation_study}
We conduct an ablation study of \deltashap{} to determine how each component contributes to its effectiveness and efficiency. \Cref{tab:ablation_study} indicates that the baseline selection with pre-processing instead of zero-filling plays a crucial role in explaining prediction evolutions for attributing features in the final time step. Attribution normalization in~\Cref{eqn:normalization} preserves feature rankings while enhancing interpretability, as showcased in~\Cref{subsec:case_study} by satisfying the efficiency property without affecting quantitative evaluation metrics. For Shapley value sampling, decreasing the number of samples ($N=1, 10$) reduces performance, while increasing to $N=100$ slightly improves the performance through a better approximation of true Shapley values. Our default setting of $N=25$ achieves the optimal balance between computational cost and effectiveness. This further corroborates the efficiency of our Shapley value sampling component, as the original Shapley value calculation requires $2^D$ samples (\ie, $N=2^D$), which incurs prohibitive computational costs.

\subsection{Qualitative Case Study} \label{subsec:case_study}
This subsection verifies whether the attributions provided by \deltashap{} align with clinical knowledge, providing interpretations that clinicians can trust. We qualitatively analyze \deltashap{} on the MIMIC-III decompensation task using LSTM as the backbone classifier~(\Cref{fig:case_1,fig:case_2,fig:case_3,fig:case_4}).
In~\Cref{fig:case_1} and~\Cref{fig:case_2}, \deltashap{} identifies oxygen saturation as a key feature for patient state prediction---capturing a drop from 92\% to 60\% in~\Cref{fig:case_1} and a recovery from 60\% to 98\% in~\Cref{fig:case_2}, both clinically significant as values below 70\% indicate acute danger. Such hypoxic states are known risk factors for ADHF~\citep{joseph2009acute}.
\Cref{fig:case_2} and~\Cref{fig:case_3} further show \deltashap{} attributing high importance to hyperglycemia exceeding 300mg/dL, consistent with evidence linking elevated glucose levels to cardiac decompensation~\citep{kattel2017association}.
\Cref{fig:case_4} highlights \deltashap{}'s detection of sharp declines in SBP and DBP as major contributors to decompensation risk, aligning with clinical understanding that such drops impair cardiac output and organ perfusion~\citep{farnett1991j}.
To summarize, these results demonstrate that \deltashap{} produces feature attributions that are not only interpretable but also clinically valid, mirroring known physiological risk factors.
Further analyses in~\Cref{fig:further_case_1,fig:further_case_2,fig:further_case_3} confirm that \deltashap{} consistently provides compact and clinically reasonable feature attributions across diverse cases, reinforcing its utility as an interpretable method aligned with clinical knowledge.
\section{Conclusion} \label{sec:conclusion}
In this paper, we proposed \deltashap{}, a novel XAI method designed to address the unique challenges of interpreting consecutive predictions in online patient monitoring within clinical settings.
By focusing on directional attributions and leveraging adapted Shapley values, our method accurately captures feature attributions while maintaining computational efficiency.
Experimental validation confirms substantial gains in both explanation fidelity and processing speed across critical care prediction tasks.
The value of our work lies in synchronizing model explanations with clinical workflows, addressing interpretability challenges in sequential medical data, and potentially enhancing physician trust in automated monitoring systems. We believe this paper advances the integration of transparent AI solutions in the healthcare domain, where interpretable, rapid insights directly impact patient outcomes.

\paragraph{Limitations.}
While \deltashap{} shows promise in explaining prediction evolutions for online patient monitoring, it has some limitations. First, by focusing on features at the last time step, it may miss delayed effects from earlier observations. Second, its applicability to fixed-point time series analysis is not yet well studied. Last, complex internal gating mechanisms may obscure the feature interactions \deltashap{} seeks to capture.

\paragraph{Future work.}
While \deltashap{} provides faithful and efficient explanations of prediction changes in online patient monitoring, our work opens broader questions about how to generalize diverse explainability techniques to this setting. In particular, most existing XAI methods are designed for static, single-time predictions and are not directly applicable to evolving prediction trajectories. A promising future direction is to systematically adapt diverse XAI methods to online time series settings.
\section*{Acknowledgements}
This work was supported by AITRICS and by the Institute for Information \& Communications Technology Planning \& Evaluation (IITP) grant funded by the Korea government (MSIP) (No. 2019-0-00075, Artificial Intelligence Graduate School Program, KAIST).

\bibliography{main/reference}
\bibliographystyle{icml2025}

\newpage
\appendix
\onecolumn
\section{Algorithm} \label{subsec:algorithm}
We provide the detailed procedure of \deltashap{} in~\Cref{alg:deltashap}, which estimates feature attributions only at the final time step for newly observed data. In detail, \deltashap{} computes prediction differences between complete and partial observations, samples permutations of feature arrival orders to approximate Shapley values for time series data, calculates marginal contributions of individual features, and normalizes the attributions to satisfy the efficiency property.

\begin{algorithm*}[!h]
\caption{\deltashap} \label{alg:deltashap}
\begin{algorithmic}[1]
\State \textbf{Input:} Time series data $\mathbf{X}_{T-W+1:T}$, Pre-trained model $f$, Number of Shapley value sampling $N$, Total feature set $\mathcal{F} =\{1, \dots, D\}$, Observed feature set $\mathcal{F}_{\text{obs}} \subset \{1, \dots, D\}$
\State $\Delta \gets f(\mathbf{X}_{T-W+1:T}) - f(\mathbf{X}_{T-W+1:T} \setminus \mathbf{X}_T)$ \Comment{\textcolor{darkgray}{Prediction difference}}
\State $\hat{\phi}_j (f, \mathbf{X}_{T-W+1:T}) \gets 0$ for all $j \in \{1, \dots, D\}$ \Comment{\textcolor{darkgray}{Initialize attributions}}
\State $\Omega \gets \emptyset$ \Comment{\textcolor{darkgray}{Initialize permutation set}}
\For{$i = 1$ to $N$} \Comment{\textcolor{darkgray}{Generate random permutations}}
    \State $\pi_i \gets$ Random permutation of all features in $\mathcal{F}_{\text{obs}}$
    \State $\Omega \gets \Omega \cup \{\pi_i\}$
\EndFor
\For{$j \in \mathcal{F}_{\text{obs}}$} \Comment{\textcolor{darkgray}{Parallelized in our implementation}}
    \For{$\pi \in \Omega$} \Comment{\textcolor{darkgray}{Also parallelized in our implementation}}
        \State $S_{\pi,j} \gets \{k \mid k \in \pi, \text{index}(k, \pi) < \text{index}(j, \pi)\}$ \Comment{\textcolor{darkgray}{Set of features preceding $j$ in permutation $\pi$}}
        
        \State $\mathbf{X}_T^{S_{\pi,j}} \gets$ Apply missing value handling to create partial observation
        \State $\mathbf{X}_T^{S_{\pi,j} \cup \{j\}} \gets \mathbf{X}_T^{S_{\pi,j}}$ but with feature $j$ observed
        
        \State $v(S_{\pi,j}) \gets f\left(\mathbf{X}_{T-W+1:T-1} \cup \mathbf{X}_T^{S_{\pi,j}}\right) - f\left(\mathbf{X}_{T-W+1:T} \setminus \mathbf{X}_T\right)$
        \State $v(S_{\pi,j} \cup \{j\}) \gets f\left(\mathbf{X}_{T-W+1:T-1} \cup \mathbf{X}_T^{S_{\pi,j} \cup \{j\}}\right) - f\left(\mathbf{X}_{T-W+1:T} \setminus \mathbf{X}_T\right)$
        \State $\hat{\phi}_j (f, \mathbf{X}_{T-W+1:T}) \gets \hat{\phi}_j (f, \mathbf{X}_{T-W+1:T}) + \frac{1}{N} \left[ v(S_{\pi,j} \cup \{j\}) - v(S_{\pi,j}) \right]$
    \EndFor
\EndFor
\State $\phi_j (f, \mathbf{X}_{T-W+1:T}) \gets \hat{\phi}_j (f, \mathbf{X}_{T-W+1:T}) \cdot \Delta / \sum_{k \in \mathcal{F}_{\text{obs}}} \hat{\phi}_k (f, \mathbf{X}_{T-W+1:T})$ for all $j \in \mathcal{F}_{\text{obs}}$ \Comment{\textcolor{darkgray}{Ensure efficiency property}}
\State \textbf{Output:} $\phi(f, \mathbf{X}_{T-W+1:T}) = \left( \phi_j(f, \mathbf{X}_{T-W+1:T}) \right)_{j=1}^{D}$
\end{algorithmic}
\end{algorithm*}
\section{Qualitative Examples}
\begin{figure*}[!t]
\centering
\includegraphics[width=\textwidth]{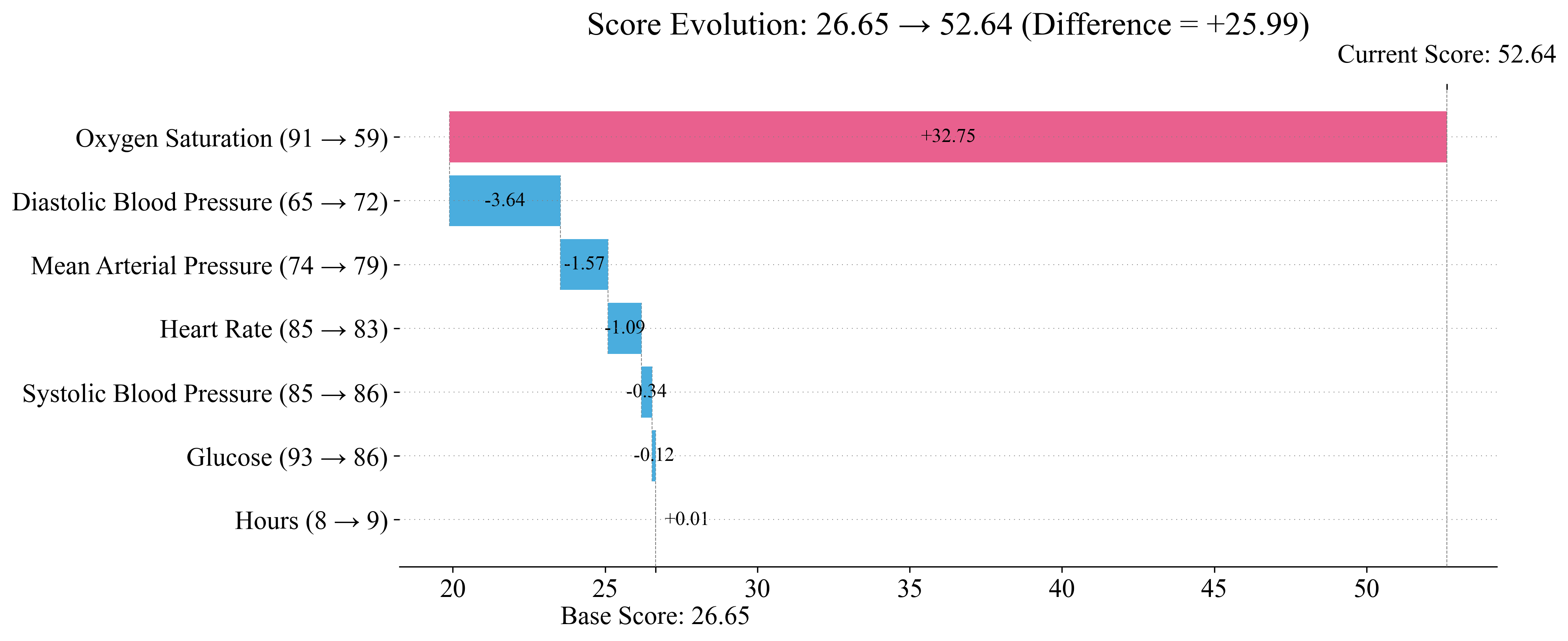}
\caption{
    Qualitative case study showing attributions extracted with \deltashap\ on MIMIC-III~\citep{mimic3} using LSTM~\citep{lstm}. This example demonstrates how decreased oxygen saturation leads to a significant increase in the decompensation risk score, illustrating \deltashap's ability to identify clinically relevant feature contributions in critical care prediction.
}
\label{fig:case_1}
\end{figure*}

\begin{figure*}[!t]
\centering
\includegraphics[width=\textwidth]{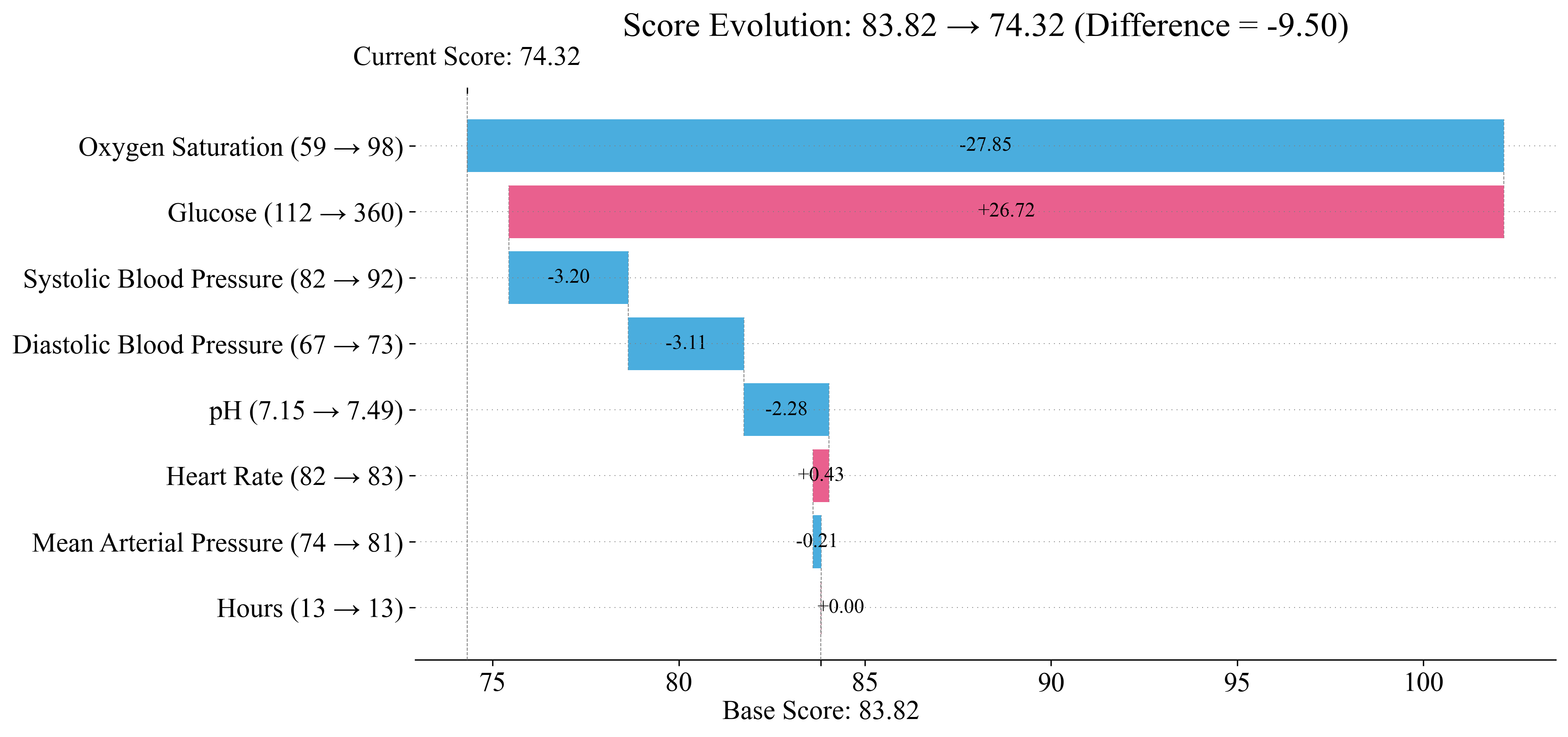}
\caption{
    Qualitative case study showing attributions extracted with \deltashap\ on MIMIC-III~\citep{mimic3} using LSTM~\citep{lstm}. This example demonstrates how elevated blood glucose levels (exceeding 300 mg/dL, where normal levels should be below 120 mg/dL) significantly contribute to increased patient risk, while increased oxygen saturation leads to a significant decrease in the decompensation risk score, highlighting \deltashap's ability to identify clinically critical deviations in physiological parameters.
}
\label{fig:case_2}
\end{figure*}

\begin{figure*}[!t]
\centering
\includegraphics[width=\textwidth]{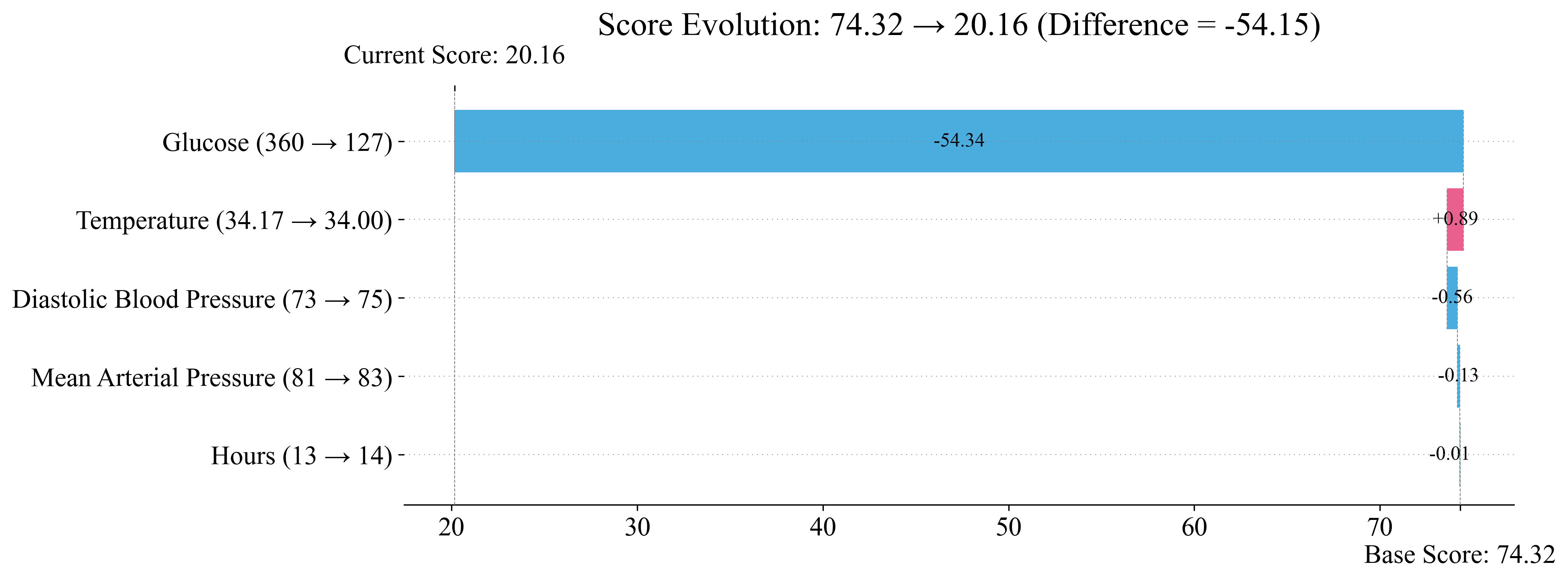}
\caption{
    Qualitative case study showing attributions extracted with \deltashap\ on MIMIC-III~\citep{mimic3} using LSTM~\citep{lstm}. This example demonstrates how a significantly decreased blood glucose level (dropping well below the normal threshold of 120 mg/dL) contributes meaningfully to patient stabilization, highlighting \deltashap's capability to identify clinically relevant physiological improvements.
}
\label{fig:case_3}
\end{figure*}

\begin{figure*}[!t]
\centering
\includegraphics[width=\textwidth]{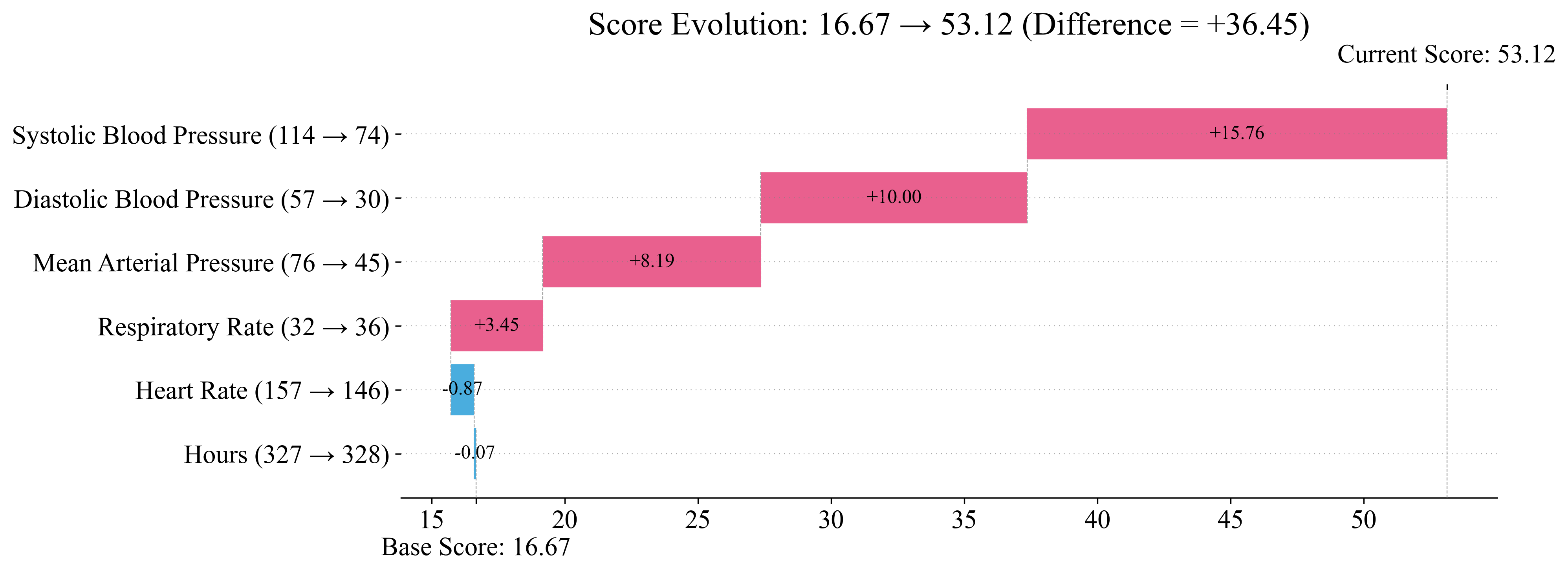}
\caption{
    Qualitative case study showing attributions extracted with \deltashap\ on MIMIC-III~\citep{mimic3} using LSTM~\citep{lstm}. This example demonstrates how a significant decrease in systolic blood pressure (SBP) and diastolic blood pressure (DBP) contributes to cardiac decompensation by reducing the heart's pumping efficiency and compromising vital organ perfusion, as accurately captured by \deltashap's attribution mechanism.
}
\label{fig:case_4}
\end{figure*}

\begin{figure*}[!t]
\centering
\includegraphics[height=.9\textheight]{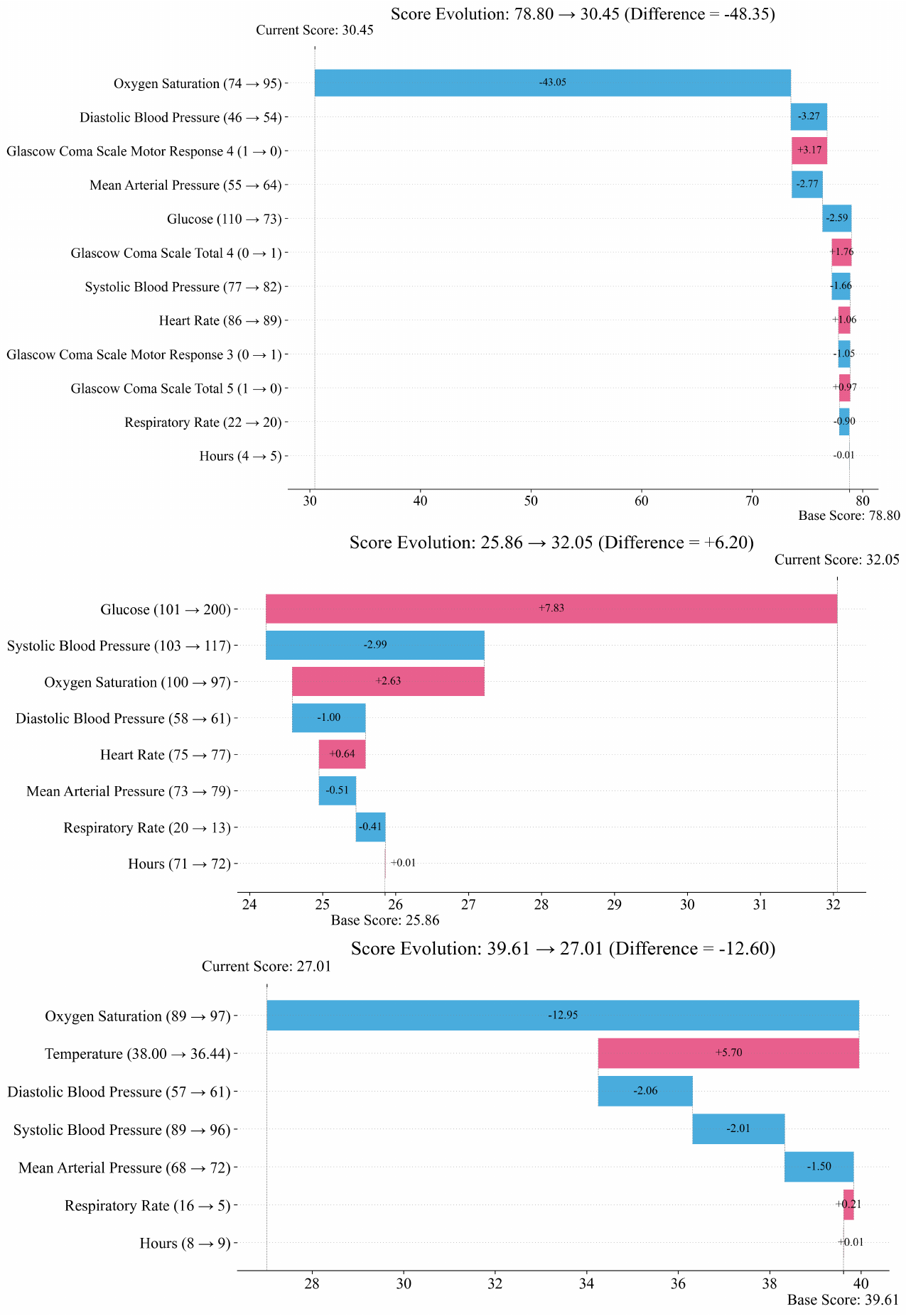}
\caption{
    Attributions extracted with \deltashap\ on MIMIC-III~\citep{mimic3} using LSTM~\citep{lstm}. These visualizations demonstrate how \deltashap\ identifies clinically relevant physiological parameters contributing to decompensation risk predictions.
}
\label{fig:further_case_1}
\end{figure*}

\begin{figure*}[!t]
\centering
\includegraphics[height=.9\textheight]{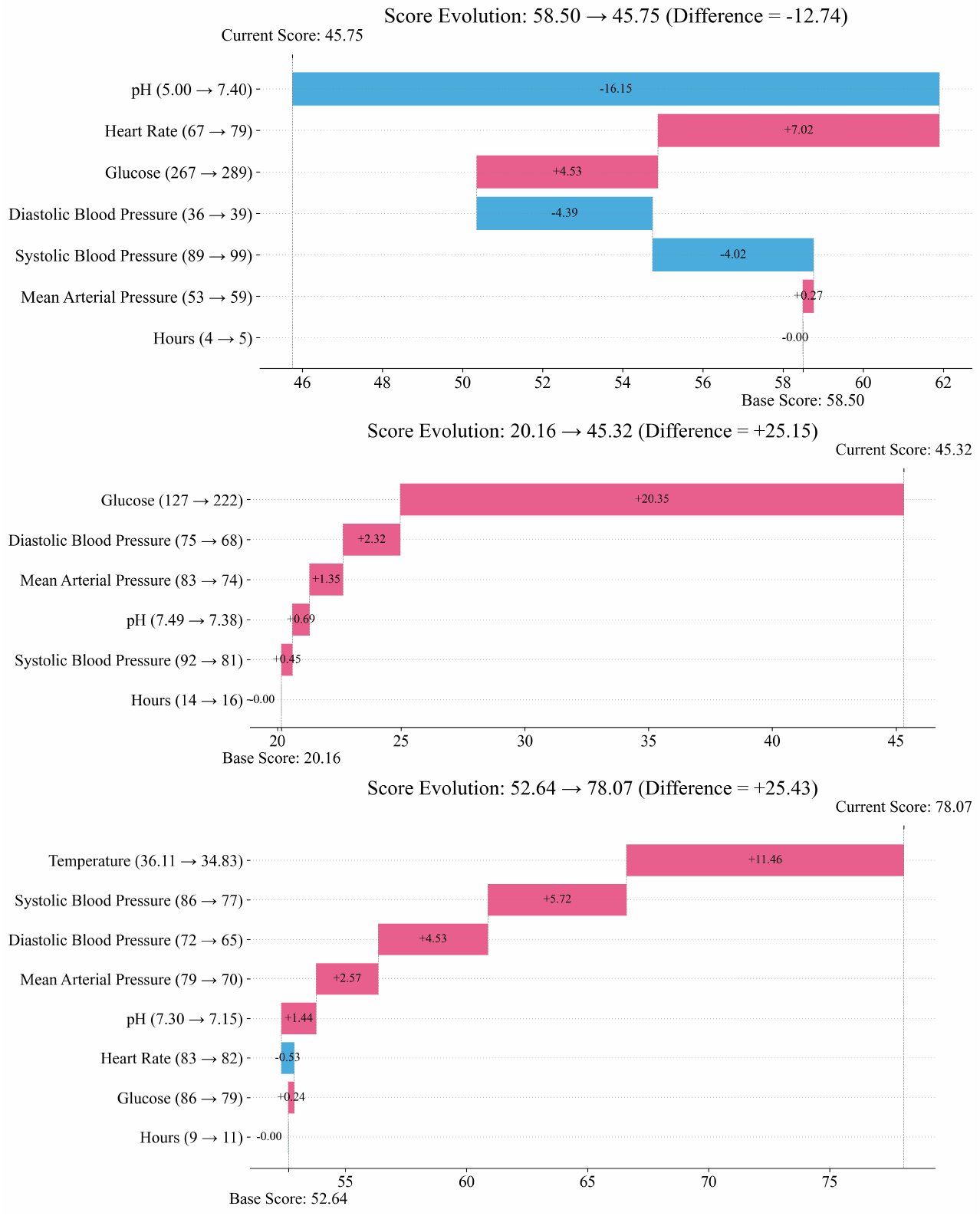}
\caption{
    Attributions extracted with \deltashap\ on MIMIC-III~\citep{mimic3} using LSTM~\citep{lstm}. These visualizations demonstrate how \deltashap\ identifies clinically relevant physiological parameters and temporal patterns in patient data that significantly influence decompensation risk predictions.
}
\label{fig:further_case_2}
\end{figure*}

\begin{figure*}[!t]
\centering
\includegraphics[height=.9\textheight]{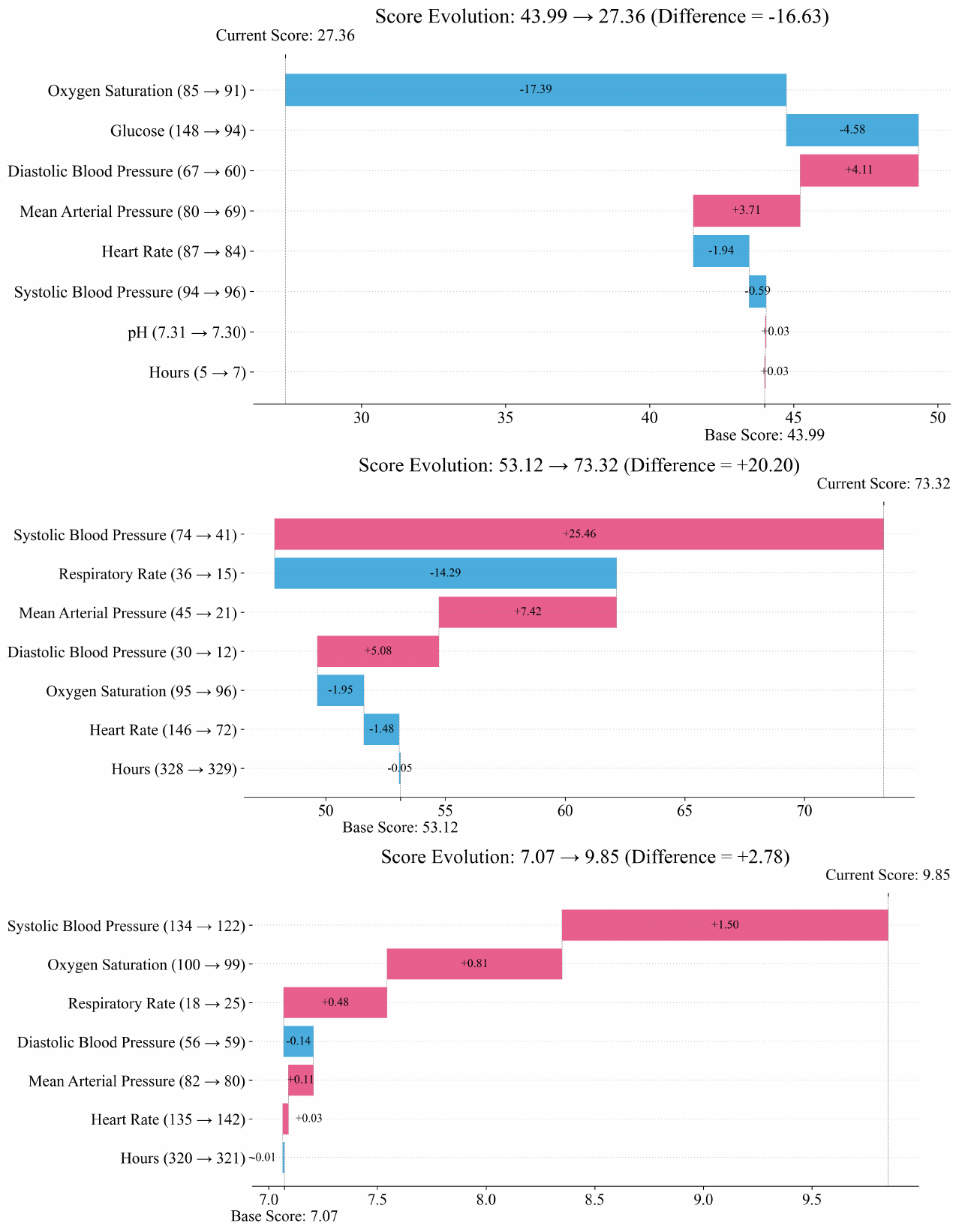}
\caption{
    Attributions extracted with \deltashap\ on MIMIC-III~\citep{mimic3} using LSTM~\citep{lstm}. These visualizations demonstrate how \deltashap\ identifies clinically relevant physiological parameters and temporal patterns in patient data that significantly influence decompensation risk predictions.
}
\label{fig:further_case_3}
\end{figure*}

We provide qualitative examples of \deltashap{}, offering detailed illustrations of its clinical relevance. The cases in \Cref{fig:case_1,fig:case_2,fig:case_3,fig:case_4} demonstrate how \deltashap{} identifies key physiological parameters that drive changes in prediction scores. Specifically, \Cref{fig:case_1,fig:case_2} visualize the impact of low oxygen saturation on the risk of decompensation, while \Cref{fig:case_2,fig:case_3} highlight the relationship between hyperglycemia and risk assessment. \Cref{fig:case_4} illustrates how hemodynamic instability influences predictions.
Additional examples in \Cref{fig:further_case_1,fig:further_case_2,fig:further_case_3} provide further validation across diverse clinical presentations, reinforcing \deltashap{}'s consistent ability to extract meaningful feature attributions that align with clinical expertise.

\clearpage


\end{document}